\documentclass[preprint,12pt]{elsarticle}

\usepackage{amssymb}
\usepackage{booktabs}
\usepackage{amsmath,amsfonts}
\usepackage{times}
\usepackage{lineno}
\usepackage{hyperref}
\usepackage{graphicx}
\usepackage{subcaption}
\usepackage{float}

\journal{arXiv}

\begin{document}

\begin{frontmatter}

\title{Empirical Ablation and Ensemble Optimization of a Convolutional Neural Network for CIFAR-10 Classification}

\author[add1]{Naser Khatti Dizabadi\corref{cor1}}
\ead{nak5300@utulsa.edu}
\cortext[cor1]{Corresponding author}

\affiliation[add1]{organization={Department of Electrical and Computer Engineering, The University of Tulsa},
            addressline={800 S. Tucker Drive},
            city={Tulsa},
            postcode={74104},
            state={OK},
            country={USA}}

\begin{abstract}
Convolutional neural networks (CNNs) remain a central approach in image classification, but their performance depends strongly on architectural and training choices. This paper presents an empirical ablation-based study of CNN optimization for the CIFAR-10 benchmark. The study evaluates 17 progressive modifications involving training duration, learning-rate scheduling, dropout configuration, pooling strategy, network depth, filter arrangement, and dense-layer design. The goal is to identify which changes improve generalization and which increase complexity without improving performance. The baseline model achieved 79.5\% test accuracy. Extending training duration improved performance steadily, whereas several structural redesigns reduced accuracy despite greater architectural variation. Based on the strongest individual configurations, a weighted ensemble was constructed, achieving 86.38\% accuracy in the reduced-data setting and 89.23\% when trained using the full CIFAR-10 dataset. These results suggest that performance gains in CNN-based classification depend less on indiscriminate increases in depth or parameter count than on careful empirical selection of training and architectural modifications. The study therefore highlights the practical value of ablation-oriented optimization and ensemble learning for small-image classification.
\end{abstract}

\begin{keyword}
Convolutional neural networks \sep CIFAR-10 \sep Image classification \sep Ablation study \sep Ensemble learning \sep Deep learning
\end{keyword}

\end{frontmatter}


\section{Introduction}
Image classification is one of the foundational tasks in computer vision, and convolutional neural networks (CNNs) have become a standard family of models for this task because they learn hierarchical visual features directly from raw image inputs. Landmark work such as AlexNet established the effectiveness of deep convolutional models for large-scale image recognition, while later architectures such as VGG and ResNet showed that depth can improve visual representation learning when supported by suitable design and optimization strategies \cite{alexnet,vgg,resnet}. At the same time, prior work on dropout and batch normalization demonstrated that training stability and regularization are often as important as raw architectural size \cite{dropout,batchnorm}.

These broader developments motivate an important practical question: when improving a CNN, which specific changes actually help, and which ones merely add complexity? This question is especially relevant on compact benchmarks such as CIFAR-10, which contains 60{,}000 color images of size \(32\times32\) across 10 classes and has long served as a standard dataset for comparing image-classification methods \cite{cifar10}.

This paper addresses that question through an empirical ablation-oriented analysis of CNN optimization for CIFAR-10 classification. Rather than proposing a fundamentally new architecture, the study starts from a baseline CNN and applies a sequence of controlled modifications involving training duration, learning-rate scheduling, dropout and pooling design, depth reduction and expansion, filter reconfiguration, and ensemble learning. The baseline model achieved 79.5\% accuracy, and the best final configuration reached 89.23\% accuracy after the strongest models were combined and evaluated with the full dataset.

The contribution of this paper is therefore empirical rather than architectural. First, it provides a structured comparison of 17 CNN modifications derived from a common baseline. Second, it shows that some seemingly stronger changes, including certain deeper or more heavily modified variants, do not reliably improve classification accuracy. Third, it demonstrates that weighted ensembling of selected high-performing models yields the strongest overall result in this study.

\section{Related Work}
CNNs became dominant in image classification largely because they can learn spatially organized feature hierarchies through repeated convolution, nonlinearity, and pooling. AlexNet showed that deep convolutional models trained with modern optimization methods could substantially improve large-scale visual recognition, and VGG later demonstrated that deeper networks built from small \(3\times3\) filters could further improve accuracy \cite{alexnet,vgg}. ResNet then addressed the difficulty of training very deep models by introducing residual learning, showing that increased depth can be beneficial when optimization barriers are reduced \cite{resnet}.

Regularization and normalization also play a major role in CNN performance. Dropout was introduced to reduce overfitting by randomly omitting units during training, and batch normalization was proposed to stabilize layer input distributions and accelerate deep-network training \cite{dropout,batchnorm}. These ideas are directly relevant to the present work because several experiments in this study test the effect of removing or modifying dropout, normalization, and related components in a baseline CNN.

CIFAR-10 has remained an important benchmark for such experiments because it is computationally manageable while still challenging enough to reveal meaningful differences among architectures and training procedures \cite{cifar10}. Another relevant area is ensemble learning. The general principle of ensembling is that combining multiple predictors can reduce error when the component models capture partly different patterns or make different mistakes \cite{bagging}. That logic remains especially useful in neural-network classification when no single model is clearly dominant but several strong variants exist.

Unlike studies whose main contribution is a new CNN architecture, this paper offers a controlled empirical study of optimization choices within a CNN framework already built around established components. Its value lies in showing which modifications helped this baseline on CIFAR-10, which did not, and how selected variants could be combined to improve final performance.

\section{Dataset and Baseline Model}

\subsection{Dataset}
The benchmark used in this study is CIFAR-10. Officially, CIFAR-10 contains 60{,}000 color images of size \(32\times32\), divided into 10 mutually exclusive classes \cite{cifar10}. In this paper, the first phase used only part of the dataset, specifically 25{,}000 images for training and 5{,}000 for testing. In the final phase, the full CIFAR-10 dataset was used to evaluate the strongest overall approach.

\subsection{Baseline CNN}
The baseline model was a modified CNN that was trained for 25 epochs and achieved 79.5\% test accuracy. The architecture contained six convolutional layers and 552{,}874 parameters. Its filter configuration was 32, 32, 64, 64, 128, and 128. The model also included seven batch-normalization layers, max-pooling after every two convolutional layers, dropout layers with rates of 0.3, 0.5, 0.5, and 0.5, and a dense layer with 128 neurons before the final classification layer. This model served as the experimental anchor for all later comparisons.

\section{Methodology}
The methodology is based on progressive empirical comparison. Instead of replacing the baseline with a single alternative network, the project introduced a series of modifications and tracked their effect on test accuracy. In total, 17 experimental steps were examined. These included increasing the number of epochs, introducing a dynamic learning-rate schedule, removing dropout and max-pooling layers, reducing and increasing model depth, changing filter counts and kernel sizes, altering dense-layer structure, removing batch normalization, and finally constructing a weighted ensemble from selected models.

The first group of experiments focused on training schedule. Because the baseline appeared undertrained at 25 epochs, the model was retrained for 50, 100, and 200 epochs. These changes produced a steady pattern of improvement, with test accuracy rising to 81.64\%, 83.36\%, and 84.60\%, respectively.

\begin{figure}[H]
\centering
\begin{subfigure}{0.48\linewidth}
\includegraphics[width=\linewidth]{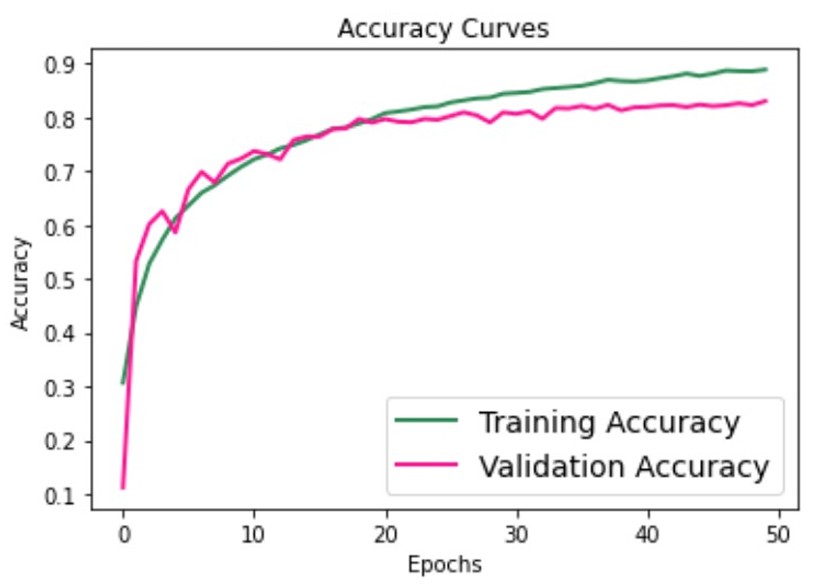}
\caption{Training and validation accuracy.}
\end{subfigure}
\hfill
\begin{subfigure}{0.48\linewidth}
\includegraphics[width=\linewidth]{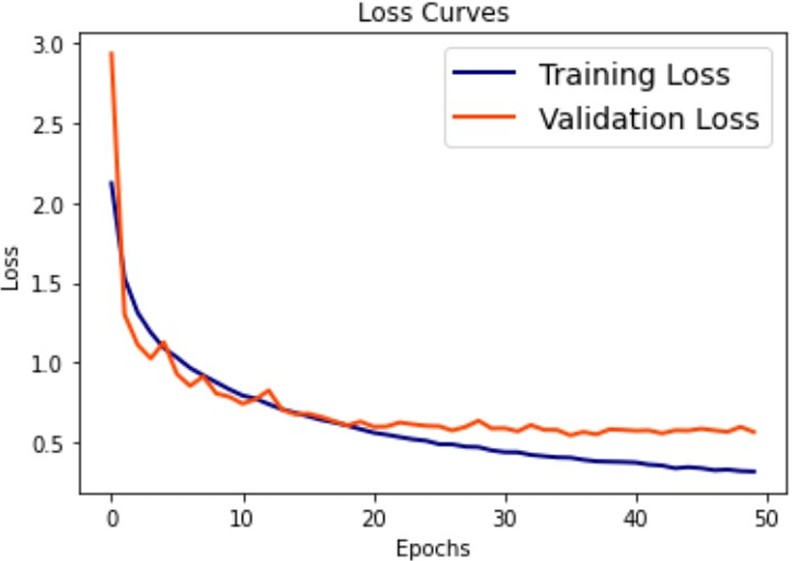}
\caption{Training and validation loss.}
\end{subfigure}
\caption{Step 1 results after retraining the baseline CNN for 50 epochs. Panel (a) shows the accuracy curves for training and validation data, and panel (b) shows the corresponding loss curves.}
\label{fig:step1}
\end{figure}

\begin{figure}[H]
\centering
\begin{subfigure}{0.48\linewidth}
\includegraphics[width=\linewidth]{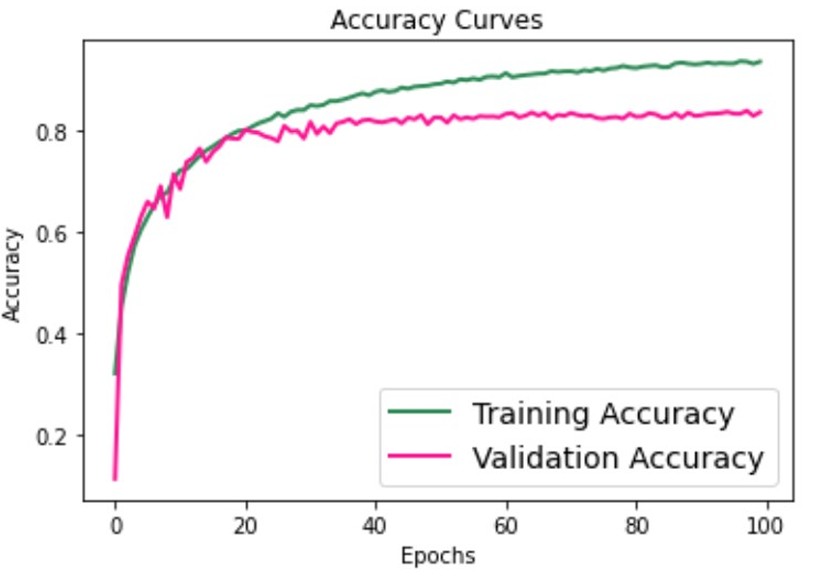}
\caption{Training and validation accuracy.}
\end{subfigure}
\hfill
\begin{subfigure}{0.48\linewidth}
\includegraphics[width=\linewidth]{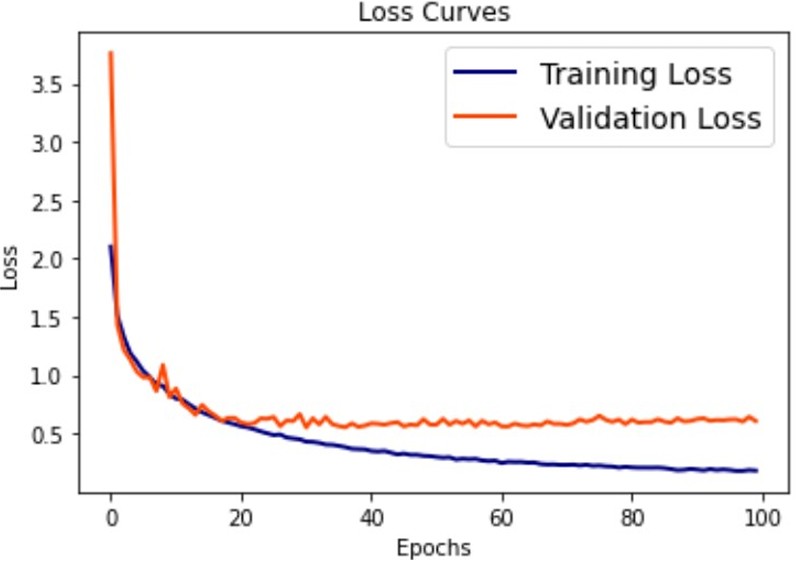}
\caption{Training and validation loss.}
\end{subfigure}
\caption{Step 2 results after increasing training to 100 epochs. Panel (a) presents the accuracy curves and panel (b) presents the loss curves for training and validation data.}
\label{fig:step2}
\end{figure}

\begin{figure}[H]
\centering
\begin{subfigure}{0.48\linewidth}
\includegraphics[width=\linewidth]{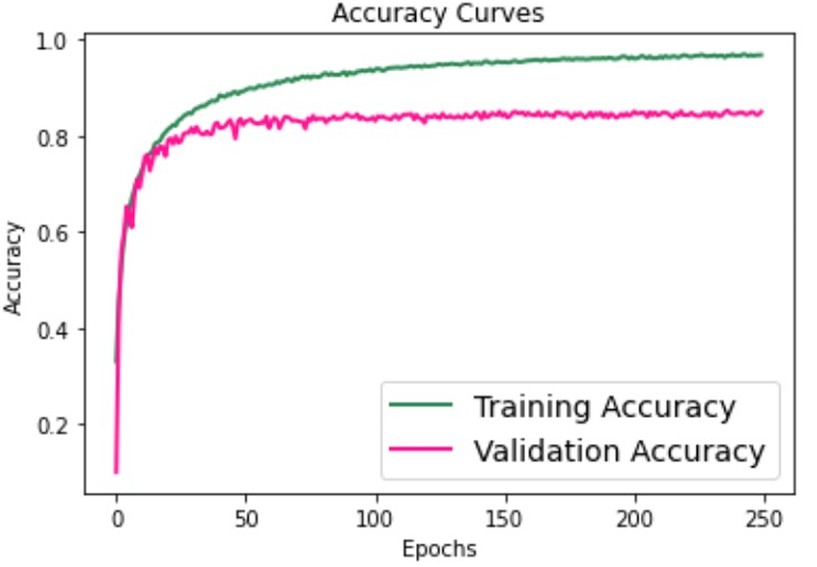}
\caption{Training and validation accuracy.}
\end{subfigure}
\hfill
\begin{subfigure}{0.48\linewidth}
\includegraphics[width=\linewidth]{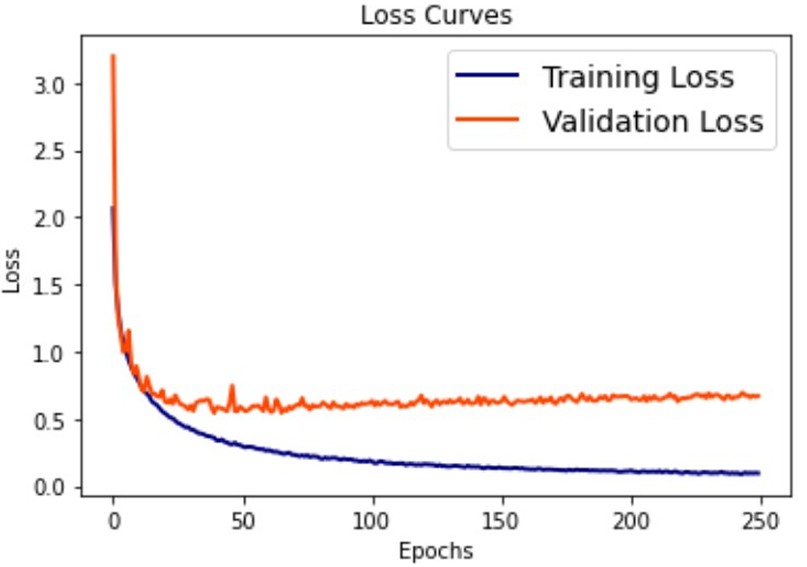}
\caption{Training and validation loss.}
\end{subfigure}
\caption{Step 3 results after training the model for 200 epochs. Panel (a) shows the accuracy curves for training and validation data, while panel (b) shows the corresponding loss curves.}
\label{fig:step3}
\end{figure}

A dynamic learning-rate schedule was then tested by training for 200 epochs at 0.001 followed by 50 epochs at 0.0001, yielding 84.28\% test accuracy.

\begin{figure}[H]
\centering
\begin{subfigure}{0.48\linewidth}
\includegraphics[width=\linewidth]{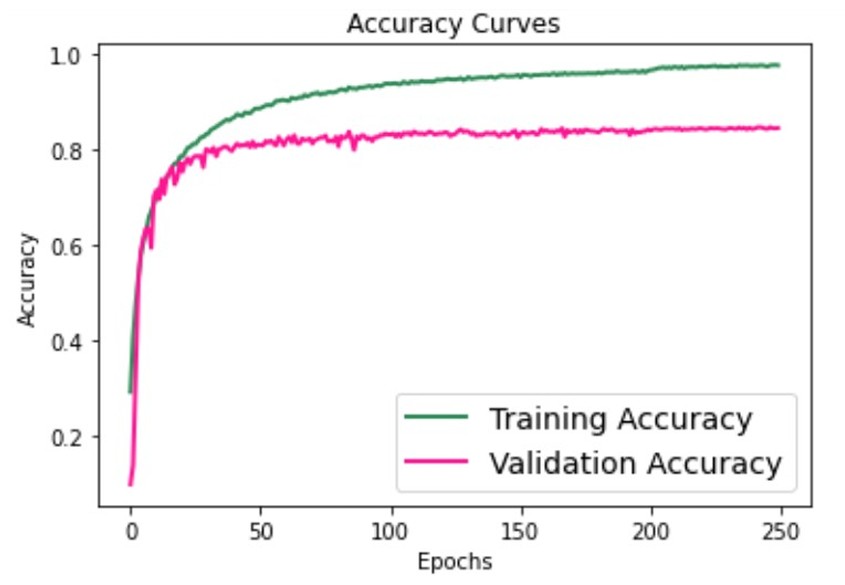}
\caption{Training and validation accuracy.}
\end{subfigure}
\hfill
\begin{subfigure}{0.48\linewidth}
\includegraphics[width=\linewidth]{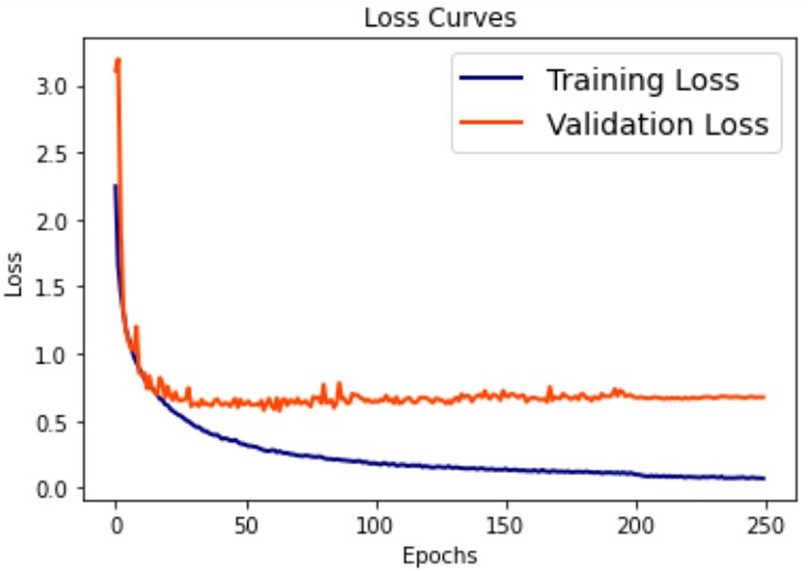}
\caption{Training and validation loss.}
\end{subfigure}
\caption{Step 4 results using a dynamic learning-rate schedule. Panel (a) reports accuracy behavior and panel (b) reports loss behavior for training and validation data.}
\label{fig:step4}
\end{figure}

The second group explored structural simplification. Removing the first three dropout layers and the last two max-pooling layers reduced performance to 74.40\%.

\begin{figure}[H]
\centering
\begin{subfigure}{0.48\linewidth}
\includegraphics[width=\linewidth]{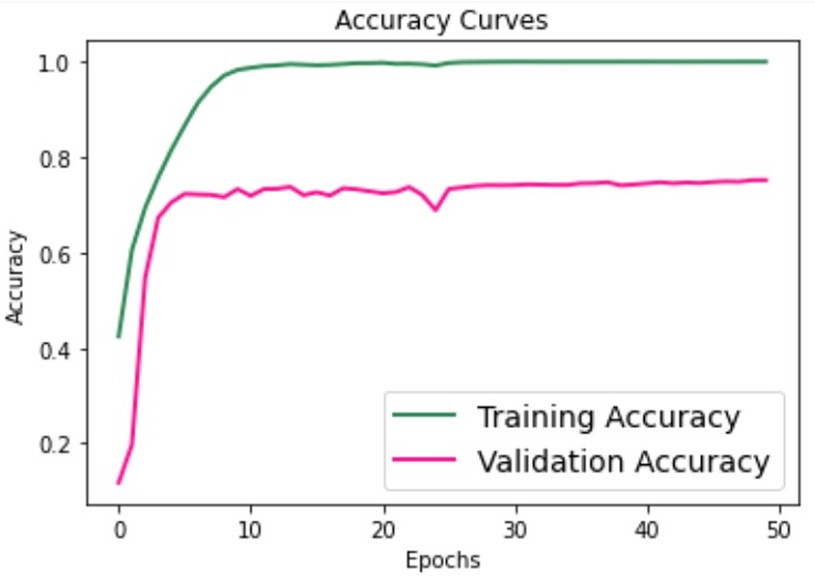}
\caption{Training and validation accuracy.}
\end{subfigure}
\hfill
\begin{subfigure}{0.48\linewidth}
\includegraphics[width=\linewidth]{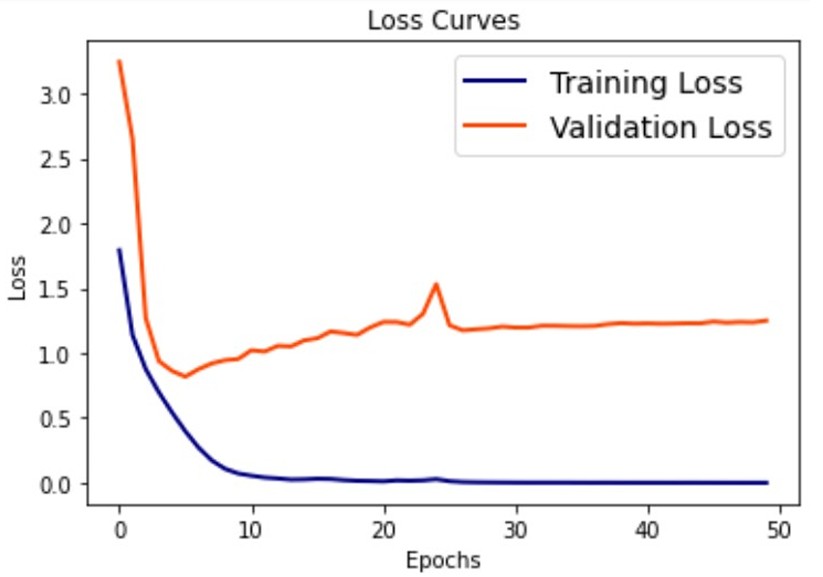}
\caption{Training and validation loss.}
\end{subfigure}
\caption{Step 5 results after removing selected dropout and max-pooling layers. Panel (a) shows the training and validation accuracy curves, and panel (b) shows the corresponding loss curves.}
\label{fig:step5}
\end{figure}

Reducing the number of convolutional layers more aggressively reduced accuracy further to 69.88\%, while a partially restored shallower variant improved only to 72.48\%.

\begin{figure}[H]
\centering
\begin{subfigure}{0.48\linewidth}
\includegraphics[width=\linewidth]{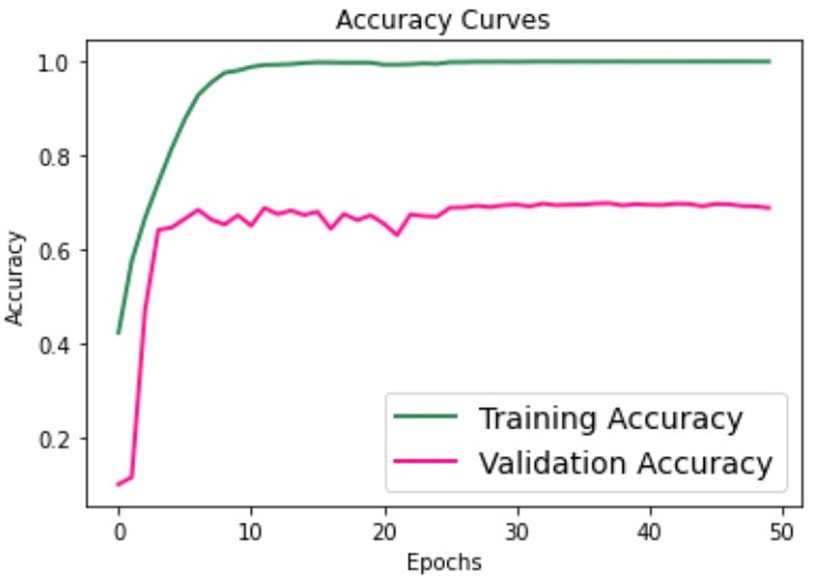}
\caption{Training and validation accuracy.}
\end{subfigure}
\hfill
\begin{subfigure}{0.48\linewidth}
\includegraphics[width=\linewidth]{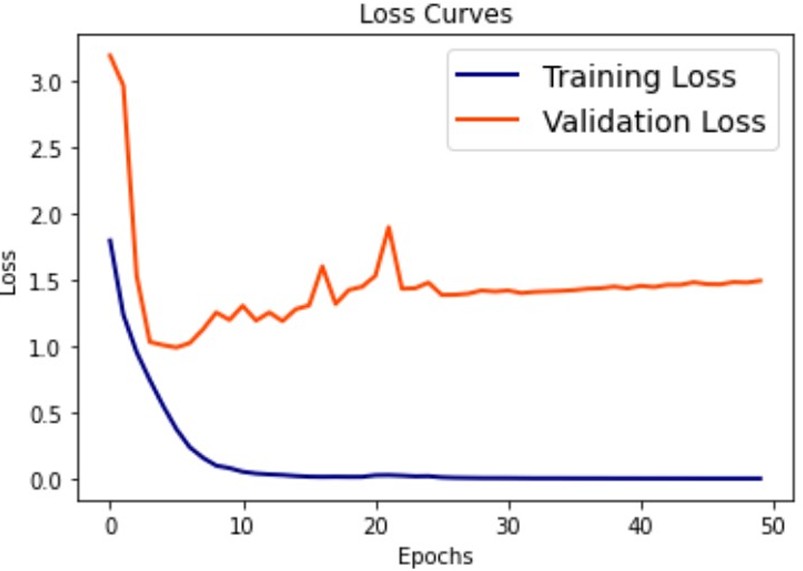}
\caption{Training and validation loss.}
\end{subfigure}
\caption{Step 6 results after reducing model depth. Panel (a) presents the accuracy curves and panel (b) presents the loss curves for training and validation data.}
\label{fig:step6}
\end{figure}

\begin{figure}[H]
\centering
\begin{subfigure}{0.48\linewidth}
\includegraphics[width=\linewidth]{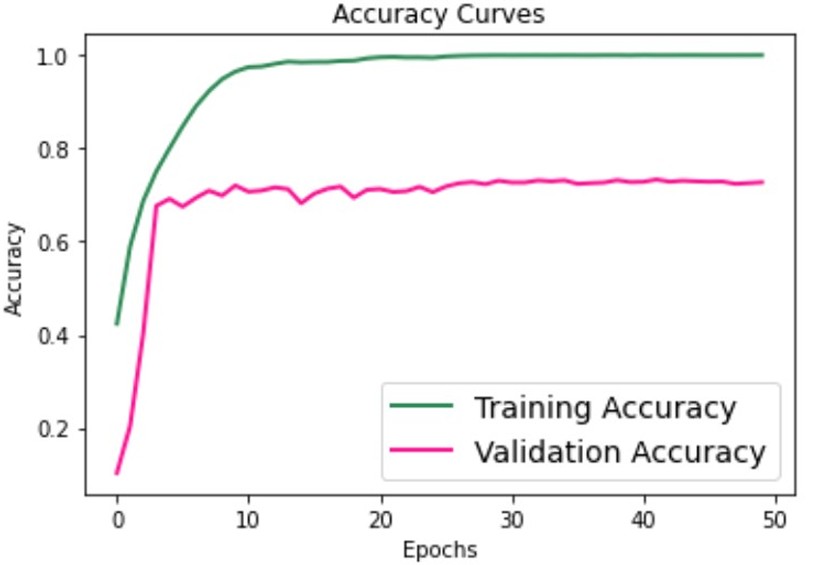}
\caption{Training and validation accuracy.}
\end{subfigure}
\hfill
\begin{subfigure}{0.48\linewidth}
\includegraphics[width=\linewidth]{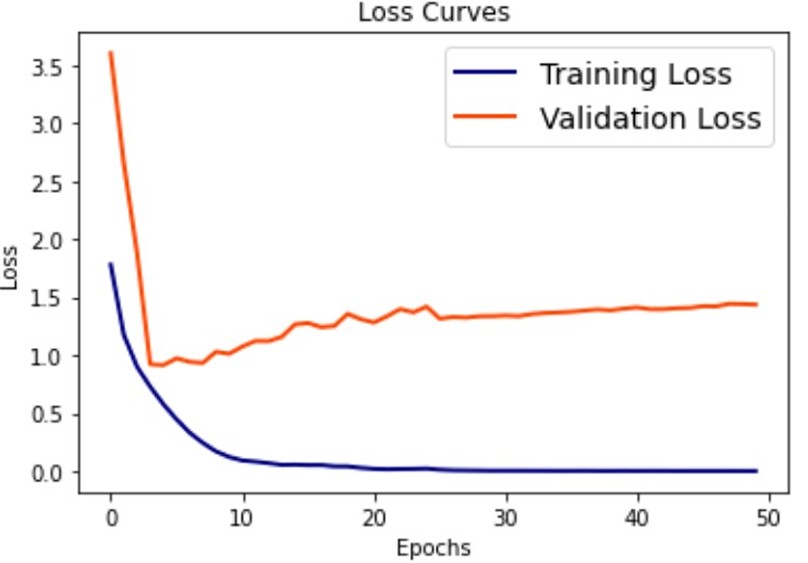}
\caption{Training and validation loss.}
\end{subfigure}
\caption{Step 7 results after partial restoration of depth and pooling. Panel (a) shows the accuracy curves and panel (b) shows the loss curves for training and validation data.}
\label{fig:step7}
\end{figure}

The third group examined capacity expansion and architectural redesign. Several deeper or wider configurations were tested, including models with more filters, changed kernel sizes, or substantially increased convolutional depth. These changes produced mixed or negative results. In one case, increasing filters and kernel-size variation resulted in 72.56\% accuracy; in another, a much deeper redesign dropped to 70.66\%.

\begin{figure}[H]
\centering
\begin{subfigure}{0.48\linewidth}
\includegraphics[width=\linewidth]{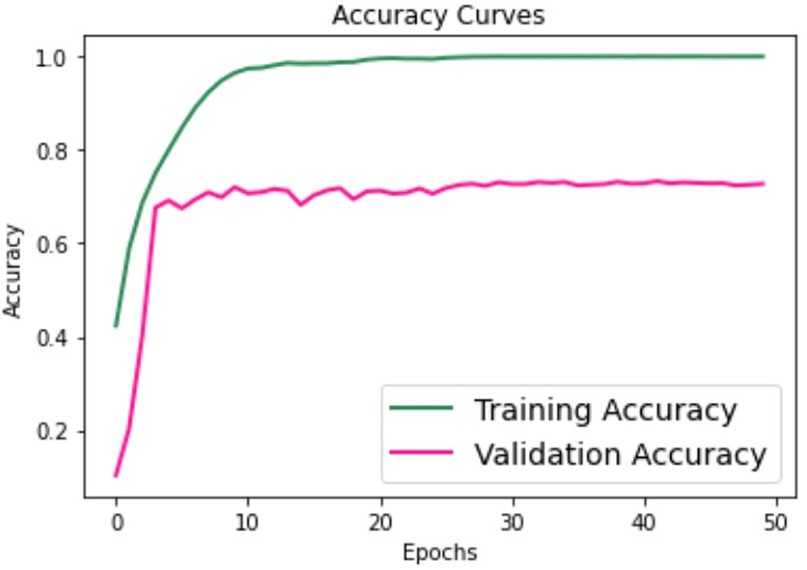}
\caption{Training and validation accuracy.}
\end{subfigure}
\hfill
\begin{subfigure}{0.48\linewidth}
\includegraphics[width=\linewidth]{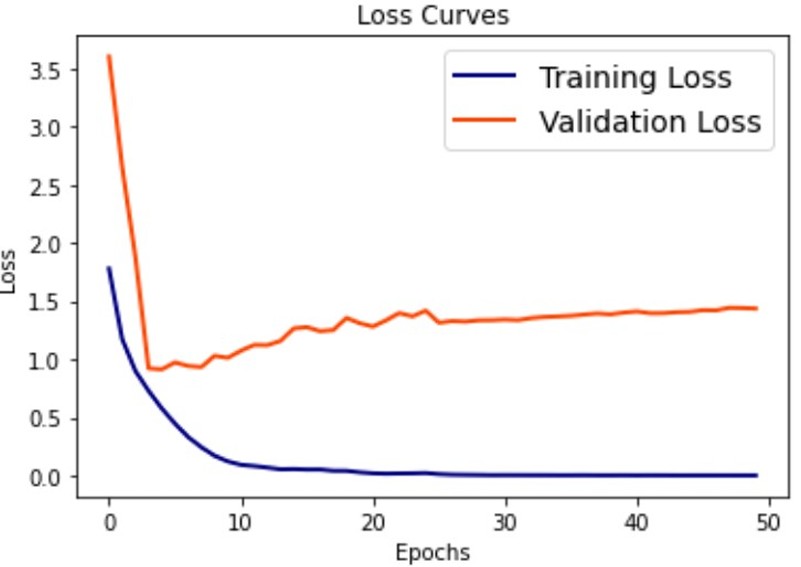}
\caption{Training and validation loss.}
\end{subfigure}
\caption{Step 8 results after increasing filters and changing kernel sizes. Panel (a) reports training and validation accuracy, while panel (b) reports training and validation loss.}
\label{fig:step8}
\end{figure}

\begin{figure}[H]
\centering
\begin{subfigure}{0.48\linewidth}
\includegraphics[width=\linewidth]{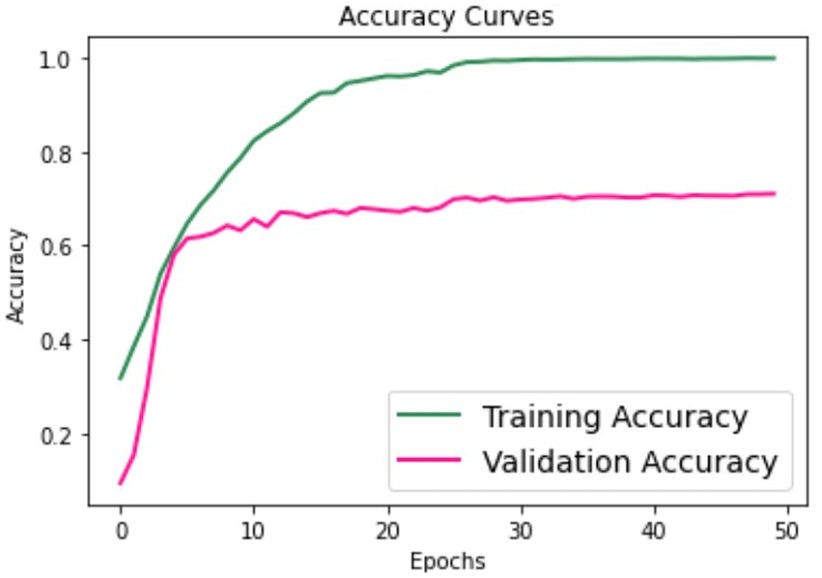}
\caption{Training and validation accuracy.}
\end{subfigure}
\hfill
\begin{subfigure}{0.48\linewidth}
\includegraphics[width=\linewidth]{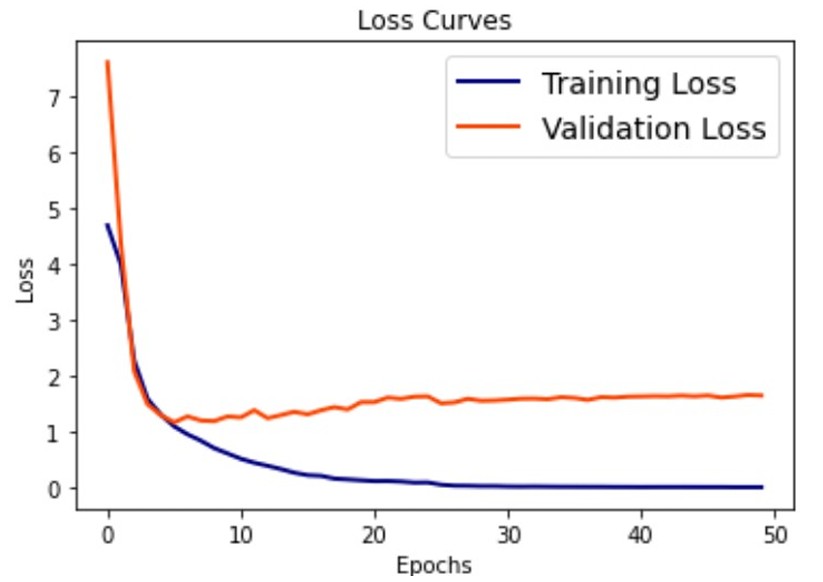}
\caption{Training and validation loss.}
\end{subfigure}
\caption{Step 9 results after a stronger increase in depth and architectural redesign. Panel (a) presents the training and validation accuracy curves, and panel (b) presents the corresponding loss curves.}
\label{fig:step9}
\end{figure}

Additional dense and pooling layers recovered some performance in later steps, reaching 80.84\% and 80.86\%, but these variants still did not clearly outperform the strongest simpler configurations.

\begin{figure}[H]
\centering
\begin{subfigure}{0.48\linewidth}
\includegraphics[width=\linewidth]{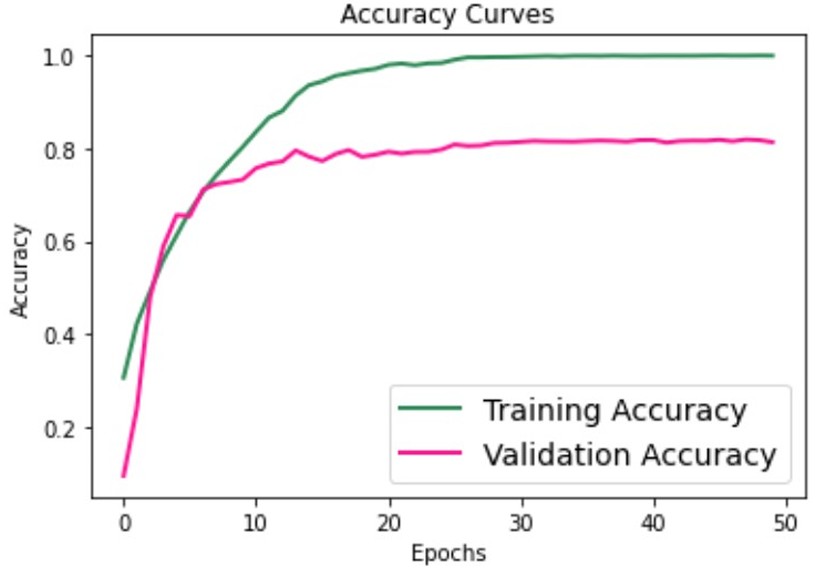}
\caption{Training and validation accuracy.}
\end{subfigure}
\hfill
\begin{subfigure}{0.48\linewidth}
\includegraphics[width=\linewidth]{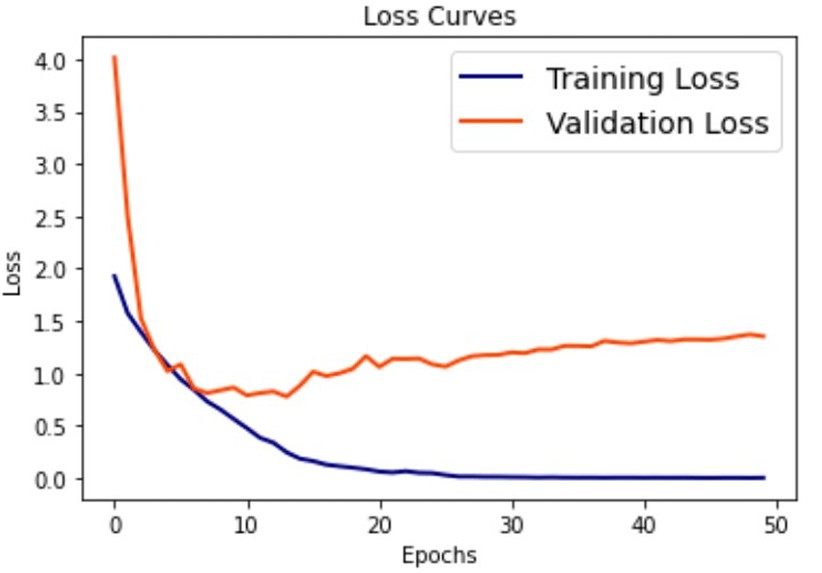}
\caption{Training and validation loss.}
\end{subfigure}
\caption{Step 10 results after adding a dense layer and max-pooling to the Step 9 model. Panel (a) shows training and validation accuracy, and panel (b) shows training and validation loss.}
\label{fig:step10}
\end{figure}

\begin{figure}[H]
\centering
\begin{subfigure}{0.48\linewidth}
\includegraphics[width=\linewidth]{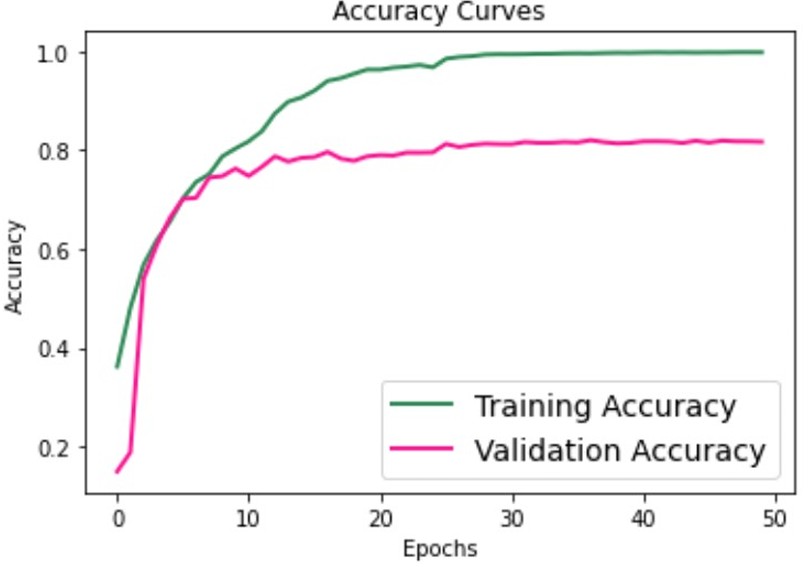}
\caption{Training and validation accuracy.}
\end{subfigure}
\hfill
\begin{subfigure}{0.48\linewidth}
\includegraphics[width=\linewidth]{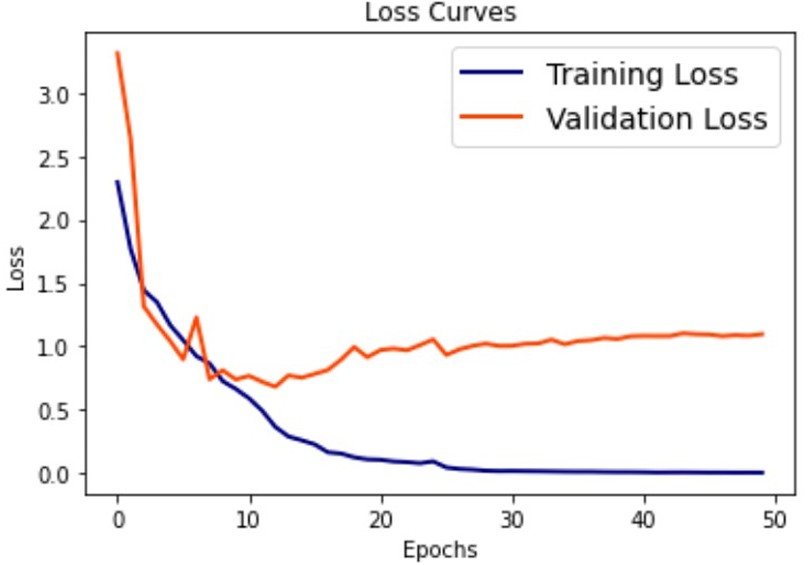}
\caption{Training and validation loss.}
\end{subfigure}
\caption{Step 11 results after adding another max-pooling layer. Panel (a) presents the training and validation accuracy curves, while panel (b) presents the loss curves.}
\label{fig:step11}
\end{figure}

Removing batch normalization, dropout, and dense layers also hurt performance, lowering accuracy to 77.56\%.

\begin{figure}[H]
\centering
\begin{subfigure}{0.48\linewidth}
\includegraphics[width=\linewidth]{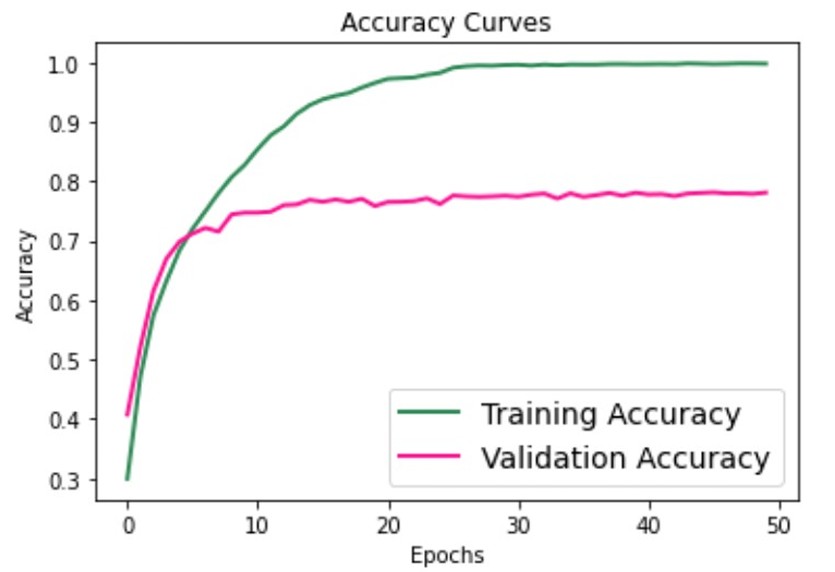}
\caption{Training and validation accuracy.}
\end{subfigure}
\hfill
\begin{subfigure}{0.48\linewidth}
\includegraphics[width=\linewidth]{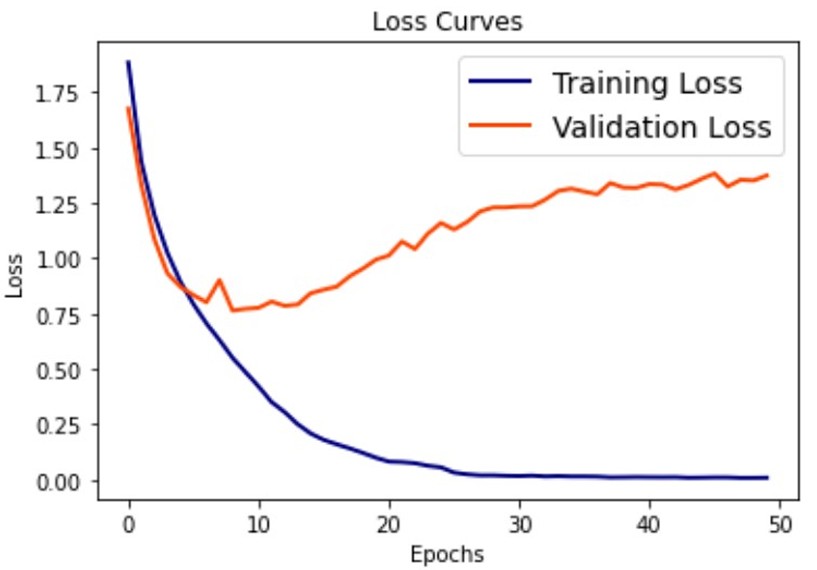}
\caption{Training and validation loss.}
\end{subfigure}
\caption{Step 12 results after removing batch normalization, dropout, and dense layers. Panel (a) shows the training and validation accuracy curves and panel (b) shows the corresponding loss curves.}
\label{fig:step12}
\end{figure}

A later baseline-style redesign with doubled filters and dynamic learning-rate scheduling produced one of the strongest individual results at 84.48\%, whereas the intermediate redesigns in Steps 13 and 14 remained weaker than the best training-schedule results.

\begin{figure}[H]
\centering
\begin{subfigure}{0.48\linewidth}
\includegraphics[width=\linewidth]{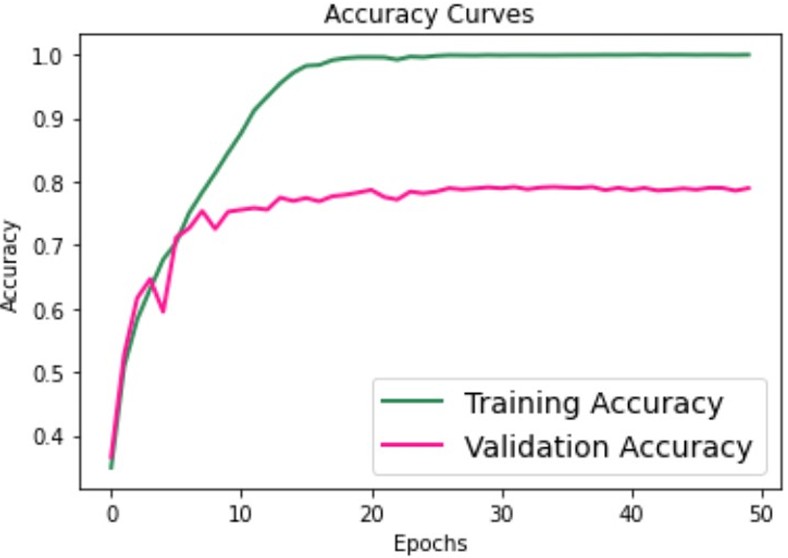}
\caption{Training and validation accuracy.}
\end{subfigure}
\hfill
\begin{subfigure}{0.48\linewidth}
\includegraphics[width=\linewidth]{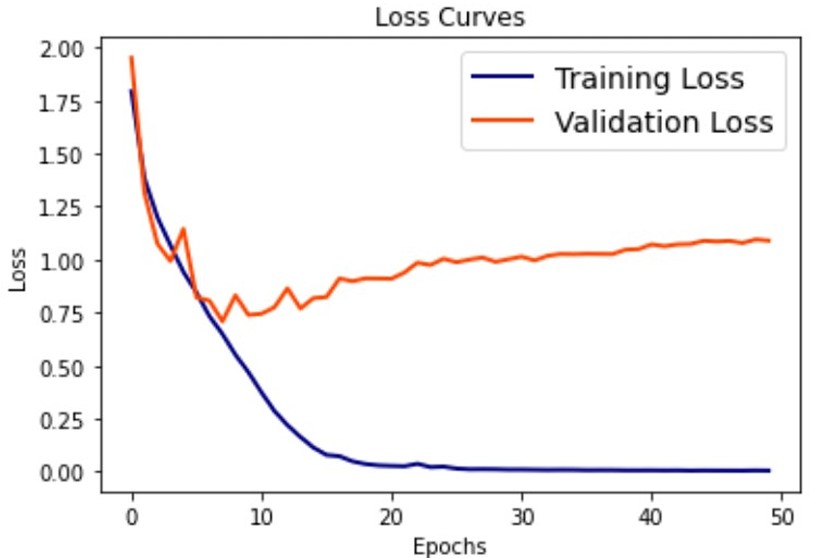}
\caption{Training and validation loss.}
\end{subfigure}
\caption{Step 13 results after doubling filters and adding a dense layer. Panel (a) presents the training and validation accuracy curves, and panel (b) presents the corresponding loss curves.}
\label{fig:step13}
\end{figure}

\begin{figure}[H]
\centering
\begin{subfigure}{0.48\linewidth}
\includegraphics[width=\linewidth]{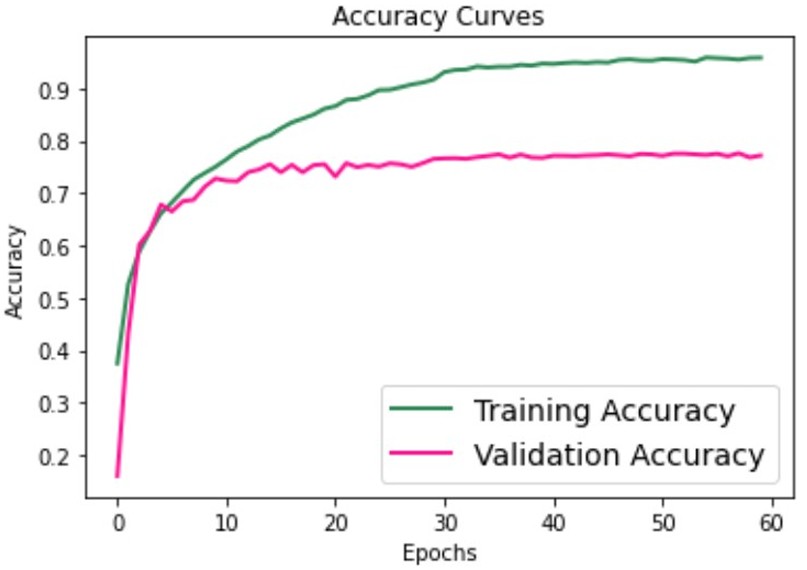}
\caption{Training and validation accuracy.}
\end{subfigure}
\hfill
\begin{subfigure}{0.48\linewidth}
\includegraphics[width=\linewidth]{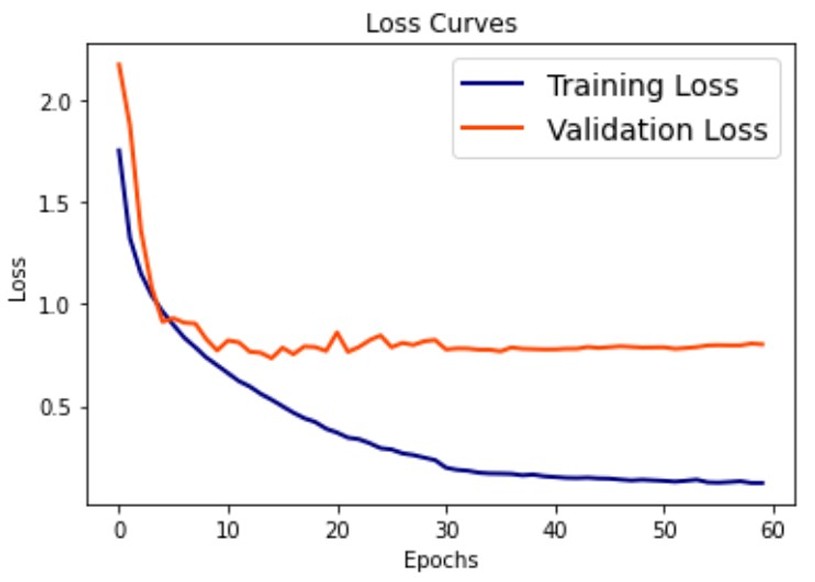}
\caption{Training and validation loss.}
\end{subfigure}
\caption{Step 14 results after reducing depth from the Step 13 model. Panel (a) reports accuracy for training and validation data, and panel (b) reports the corresponding loss.}
\label{fig:step14}
\end{figure}

\begin{figure}[H]
\centering
\begin{subfigure}{0.48\linewidth}
\includegraphics[width=\linewidth]{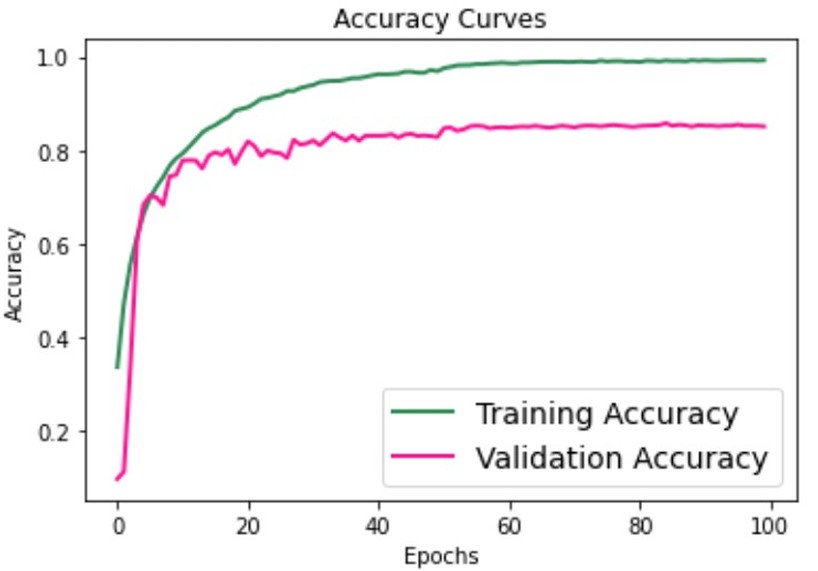}
\caption{Training and validation accuracy.}
\end{subfigure}
\hfill
\begin{subfigure}{0.48\linewidth}
\includegraphics[width=\linewidth]{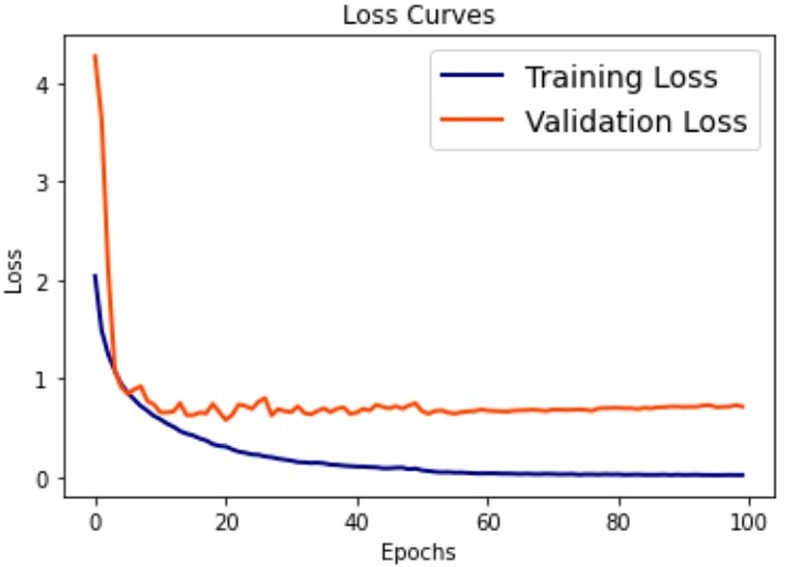}
\caption{Training and validation loss.}
\end{subfigure}
\caption{Step 15 results with doubled baseline-style filters and dynamic learning-rate scheduling. Panel (a) shows the training and validation accuracy curves, and panel (b) shows the corresponding loss curves.}
\label{fig:step15}
\end{figure}

\section{Results}

\subsection{Summary of Experimental Results}
Table~\ref{tab:results} summarizes the major experimental steps and their corresponding test accuracies.

\begin{table}[H]
\centering
\caption{Summary of CNN modifications and test accuracy}
\label{tab:results}
\begin{tabular}{@{}clc@{}}
\toprule
Step & Main modification & Test accuracy (\%) \\ \midrule
Baseline & Original CNN, 25 epochs & 79.50 \\
1 & Train for 50 epochs & 81.64 \\
2 & Train for 100 epochs & 83.36 \\
3 & Train for 200 epochs & 84.60 \\
4 & Dynamic learning-rate schedule & 84.28 \\
5 & Remove several dropout and max-pooling layers & 74.40 \\
6 & Reduce model depth & 69.88 \\
7 & Partial restoration of depth and pooling & 72.48 \\
8 & Increase filters and change kernel sizes & 72.56 \\
9 & Strong increase in depth and architectural redesign & 70.66 \\
10 & Add dense and pooling layers to Step 9 model & 80.84 \\
11 & Add another pooling layer & 80.86 \\
12 & Remove batch normalization, dropout, and dense layers & 77.56 \\
13 & Double filters and add dense layer & 78.96 \\
14 & Reduce depth of Step 13 model & 76.72 \\
15 & Double filters in baseline-style model + dynamic LR & 84.48 \\
16 & Weighted ensemble of Steps 3, 13, and 15 & 86.38 \\
17 & Ensemble using full CIFAR-10 dataset & 89.23 \\ \bottomrule
\end{tabular}
\end{table}

\subsection{Main Performance Patterns}
The most consistent improvement in the study came from extending training duration. Accuracy rose from 79.5\% in the baseline to 81.64\%, 83.36\%, and 84.60\% as the number of epochs increased from 25 to 50, 100, and 200. This pattern strongly suggests that the baseline network had not converged adequately under the original training schedule.

By contrast, many structural modifications reduced performance. Removing multiple dropout and pooling layers caused a substantial loss in accuracy, and reducing model depth produced an even larger decline. These outcomes are consistent with the broader literature showing that regularization and well-chosen feature-hierarchy design are important for stable CNN training and generalization \cite{dropout,batchnorm}.

The experiments also showed that simply increasing architectural complexity was not sufficient. Several deeper or wider networks performed worse than the baseline or worse than the strongest longer-trained variants. This result aligns with prior work suggesting that depth can be beneficial, but only when training and architecture are properly aligned; additional layers or filters do not automatically produce better generalization \cite{vgg,resnet}.

\subsection{Ensemble Performance}
The strongest overall result came from ensembling. Although no single model dominated every other variant by a large margin, the weighted ensemble of selected models achieved 86.38\% in the reduced-data setting and 89.23\% with the full dataset.

\begin{figure}[H]
\centering
\includegraphics[width=0.78\linewidth]{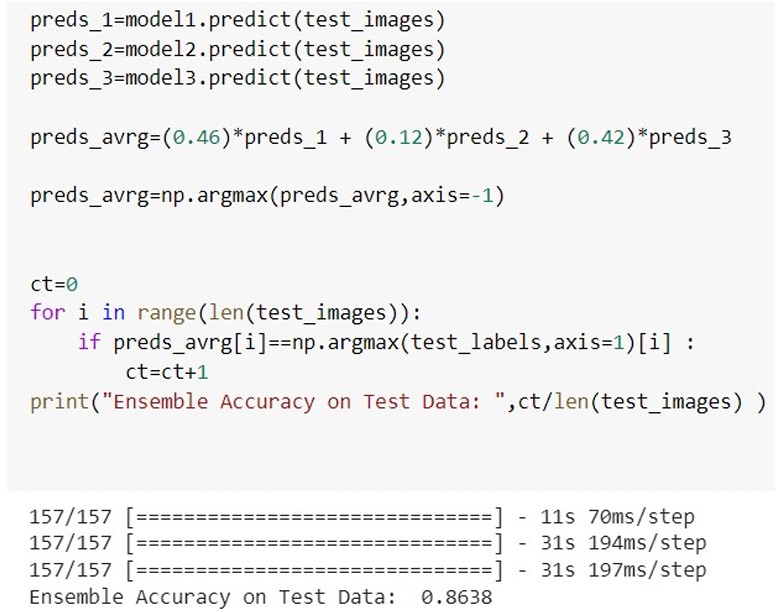}
\caption{Step 16 ensemble result showing the weighted combination of selected models and the resulting ensemble accuracy on the reduced-data test setting.}
\label{fig:step16}
\end{figure}

This suggests that the component models captured partly complementary prediction patterns and that their combination reduced the weaknesses of relying on one architecture alone. The report also included class-wise accuracy matrices for both the reduced-data and full-data settings.

\begin{figure}[H]
\centering
\includegraphics[width=0.78\linewidth]{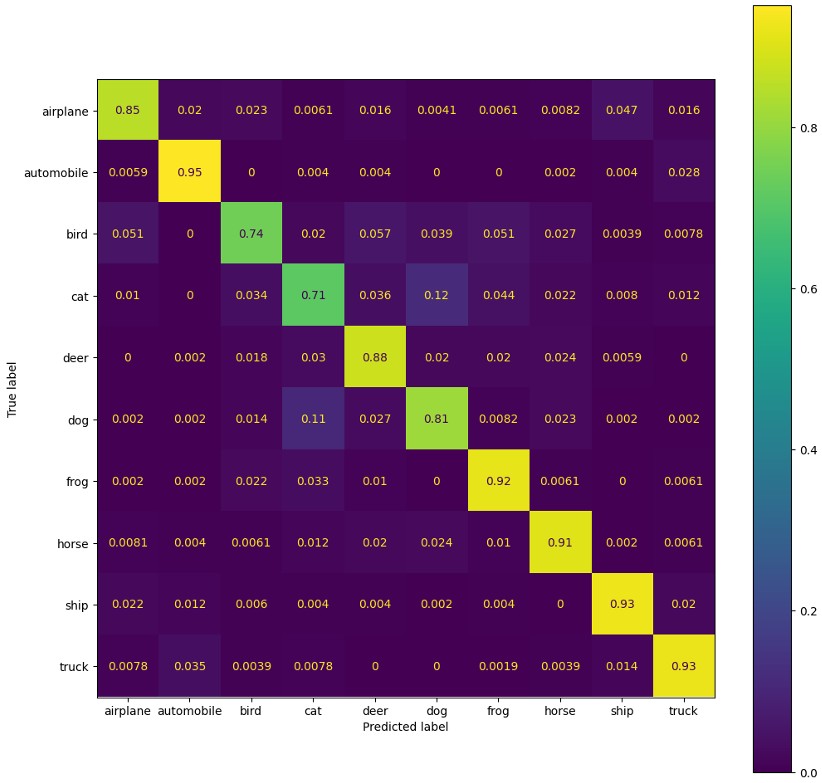}
\caption{Class-wise accuracy matrix for the reduced-data ensemble setting. The diagonal values summarize per-class recognition performance after ensembling.}
\label{fig:step17}
\end{figure}

The final experiment used the full dataset and raised the reported ensemble accuracy to 89.23\%.

\begin{figure}[H]
\centering
\includegraphics[width=0.78\linewidth]{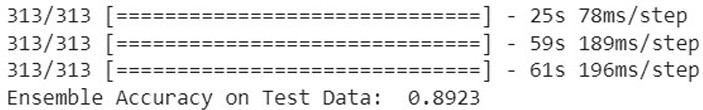}
\caption{Step 17 final result using the full CIFAR-10 dataset, reporting the ensemble accuracy on the full-data test setting.}
\label{fig:step18}
\end{figure}

The final class-wise matrix shows that performance varied across categories rather than remaining uniform.

\begin{figure}[H]
\centering
\includegraphics[width=0.78\linewidth]{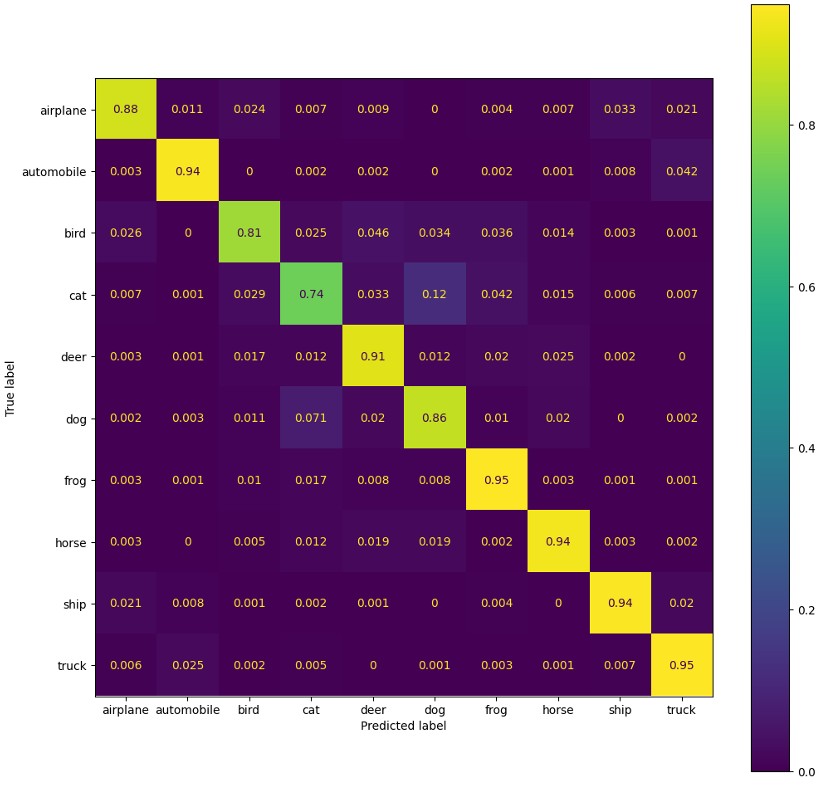}
\caption{Class-wise accuracy matrix for the final full-dataset model. The diagonal entries indicate per-class accuracy after training and evaluation on the full-data setting.}
\label{fig:step19}
\end{figure}

\section{Discussion}
The results support a restrained but important conclusion: in this project, CNN improvement came more reliably from selective empirical optimization than from indiscriminate complexity increases. Longer training was one of the most effective interventions, which suggests that better use of the original architecture mattered at least as much as architectural redesign.

The negative results are equally informative. Removing dropout, pooling, batch normalization, or other stabilizing components usually reduced performance, indicating that these layers were serving meaningful functions rather than merely inflating the architecture. Likewise, deeper or more aggressively redesigned models often failed to improve results. In practical terms, the study shows that optimization decisions should be justified experimentally rather than assumed beneficial on intuition alone.

The ensemble result is especially significant because it shows how multiple moderately strong variants can be combined into a stronger final predictor. This does not make the paper an architectural breakthrough, but it does make it a useful optimization study: the work documents a process of identifying productive modifications, rejecting unhelpful ones, and integrating the most effective models into a final system with clearly improved performance over the baseline.

\section{Limitations}
This study has several limitations. First, it evaluates only CIFAR-10, so its findings should not be assumed to generalize directly to larger or more complex datasets. Second, the report presents single-result comparisons for each configuration and does not provide repeated trials, standard deviations, or statistical significance testing. Third, the paper does not benchmark the reported models against stronger contemporary CIFAR-style architectures such as ResNet variants, which limits claims about competitiveness beyond the internal comparisons in this study. Finally, the contribution is best understood as empirical optimization of a known CNN framework rather than the proposal of a fundamentally new model.

\section{Conclusion}
This paper presented an empirical ablation-based study of CNN optimization for CIFAR-10 classification. Starting from a baseline CNN with 79.5\% accuracy, the study evaluated 17 progressive modifications involving training duration, learning-rate scheduling, regularization changes, depth modification, filter redesign, and ensemble construction. The results showed that not all changes improved performance: several deeper or more heavily altered configurations underperformed, while longer training and selected refinements produced clearer gains. The best final outcome came from a weighted ensemble, which achieved 86.38\% accuracy in the reduced-data setting and 89.23\% with the full dataset. Overall, the study shows that systematic empirical comparison can be a practical way to improve CNN performance, even when the goal is refinement rather than architectural invention.

\bibliographystyle{elsarticle-num}
\bibliography{MyBibDatabase}

\end{document}